%% file: main_arXiv.tex
\documentclass[11pt]{article}
\usepackage{fullpage}

\usepackage{booktabs}
\usepackage{color}

\usepackage{algorithm}
\usepackage{algorithmic}

\usepackage{url}

\usepackage{amssymb}
\usepackage{amsmath}
\usepackage{mathtools}
\usepackage{multicol}
\usepackage{multirow}

\newcommand{\trace}{{\rm trace}}

\newcommand*\samethanks[1][\value{footnote}]{\footnotemark[#1]}

\def\R{\mathbb{R}}

\def\calX{\mathcal{X}}
\def\calY{\mathcal{Y}}

\def\calC{\mathcal{C}}

\def\bx{{\mathbf x}}
\def\by{{\mathbf y}}

\def\grad{\mathop{\rm grad}\nolimits}

\def\bzero{{\mathbf 0}}
\def\bone{{\mathbf 1}}

\def\bx{{\mathbf x}}
\def\by{{\mathbf y}}
\def\bz{{\mathbf z}}

\def\bA{\mathbf A}

\def\bC{\mathbf C}

\def\bS{\mathbf S}

\def\bX{{\mathbf X}}

\def\bZ{{\mathbf Z}}

\def\bGamma{{\mathbf{\Gamma}}}

\def\maxop{\mathop{\rm max}\limits} 
\def\minop{\mathop{\rm min}\limits}

\renewcommand{\trace}{{\rm trace}}

\DeclareMathAlphabet\mathbfcal{OMS}{cmsy}{b}{n}

\begin{document}
\title{\LARGE \bf
Manifold optimization\\ for non-linear optimal transport problems
}

\author{Bamdev Mishra\thanks{Microsoft, India. Email: \{bamdevm, pratik.jawanpuria\}@microsoft.com.}\quad  N. T. V. Satyadev\thanks{Vayve Technologies, India. Email: tvsatyadev@gmail.com.}\quad  Hiroyuki Kasai\thanks{Waseda University, Japan. Email: hiroyuki.kasai@waseda.jp.}\quad  Pratik Jawanpuria\samethanks[1]}


\date{}

\maketitle

\input{abstract}


\input{manuscript}

\bibliographystyle{amsalpha}
\bibliography{OTreferences}

\end{document}

%% file: abstract.tex
\begin{abstract}
Optimal transport (OT) has recently found widespread interest in machine learning. It allows to define novel distances between probability measures, which have shown promise in several applications. In this work, we discuss how to computationally approach general non-linear OT problems within the framework of Riemannian manifold optimization. The basis of this is the manifold of doubly stochastic matrices (and their generalization). Even though the manifold geometry is not new, surprisingly, its usefulness for solving general non-linear OT problems has not been popular. To this end, we specifically discuss optimization-related ingredients that allow modeling the OT problem on smooth Riemannian manifolds by exploiting the geometry of the search space. We also discuss extensions where we reuse the developed optimization ingredients. We make available the {\bf M}anifold optimization-based {\bf O}ptimal {\bf T}ransport, or MOT, repository with codes useful in solving OT problems in Python and Matlab. The codes are available at \url{https://github.com/SatyadevNtv/MOT}.

\end{abstract}

%% file: manuscript.tex
\section{Introduction} 


Given two probability measures $\mu_1$ and $\mu_2$ over metric spaces $\calX$ and $\calY$, respectively, the optimal transport problem is \cite{kantorovich42a}
\begin{equation}\label{eqn:optimal-transport}
    {\rm W}_c(\mu_1,\mu_2) = \inf_{\bGamma\in\Pi(\mu_1,\mu_2)} \int_{\calX\times\calY} c(\bx,\by) d\bGamma(\bx,\by),
\end{equation}
where $\bGamma$ is the {transport plan}, $\Pi(\mu_1,\mu_2)$ is the set of joint distributions with marginals $\mu_1$ and $\mu_2$, and $c:\calX \times \calY \rightarrow \R_{+}:(\bx,\by)\rightarrow c(\bx,\by)$ represents the transportation cost function. The cost ${\rm W}_c(\mu_1,\mu_2) $ is a valid distance metric between between the probability measures $\mu_1$ and $\mu_2$ and is popularly known as the \textit{Wasserstein} distance or the earth mover's distance \cite{rubner00}. It defines a geometry over the space of probability measures and evaluates the minimal amount of work required to transform one measure into another with respect to the ground cost function \cite{villani09a}.

The optimal transport (OT) toolbox has found widespread application in domain adaptation \cite{courty17a,courty17b,titouan2020co}, computer vision \cite{rubner00,deshpande19a}, natural language processing \cite{melis18a,mjaw20a}, \cite{courty17a,nath20a}, multi-task learning \cite{janati19a}, classification \cite{frogner15a}, and generative model training \cite{genevay18a,arjovsky17a}, among others. Availability of libraries such as Python Optimal Transport (POT)~\cite{flamary2021pot} has allowed easy implementation of OT-based solutions in real-world problems. 
There have been many works in computational approaches for solving optimal transport problems \cite{cuturi13a,ferradans14a,peyre19a}. In particular, \cite{cuturi13a} has shown that the OT problem can be solved efficiently via the Sinkhorn algorithm~\cite{knight2008sinkhorn} if entropy regularization is used on the transport plan. 


In this work, we solve OT problems such as (\ref{eqn:optimal-transport}) and its non-linear generalizations \cite{peyre16a,paty19a,paty20a} by exploiting the geometry of the search space of joint distributions $\bGamma$. We view the problem (\ref{eqn:optimal-transport}) as an optimization problem on a generalized doubly stochastic manifold \cite{douik2019manifold,shi19a}. The manifold optimization framework conceptually translates a constrained optimization problem into an {unconstrained} optimization problem over the manifold. It provides systematic ways to build first and second order algorithms, e.g., the conjugate gradients and trust-regions algorithms \cite{absil2009optimization,boumal2020intromanifolds}. There exist open source tools like Pymanopt \cite{townsend2016pymanopt}, Manopt \cite{boumal2014manopt}, McTorch \cite{Meghwanshi_arXiv_2018}, ROPTLIB \cite{huang16a}, and Manopt.jl \cite{bergmann19a} which allow rapid prototyping of manifold optimization algorithms.

The main aim of the work is two fold. First, to explore the connection between manifold optimization framework and optimal transport problems. Second, to make available the {\bf M}anifold optimization based {\bf O}ptimal {\bf T}ransport, or {\bf MOT}, repository. We provide both Python and Matlab codes that integrate well with Pymanopt and Manopt toolboxes. This, we believe, would be useful for machine learning practitioners to experiment with manifold techniques for different OT problems. Conceptually, the MOT viewpoint allows a principled way to use the efficient Sinkhorn algorithm iterations (shown later) to solve various non-linear OT problems but without the constraint of employing entropic regularization on the transport plan.

The organization of the paper is as follows. In Section \ref{sec:motivation}, we discuss different optimal transport problem formulations and how they can all be seen as minimization of a function over the manifold $\Pi(\mu_1,\mu_2)$. We discus manifold optimization framework and show optimization-related ingredients in Section 
\ref{sec:manifold_optimization}. Most of Section \ref{sec:manifold_optimization} has been borrowed from \cite{douik2019manifold,shi19a}. Section \ref{sec:extension} discusses how to extend the optimization strategy to different other scenarios. In Section \ref{sec:experiments}, we show the usefulness and generality of MOT in some applications.

We make available the {\bf M}anifold optimization based {\bf O}ptimal {\bf T}ransport, or MOT, repository with codes available at \url{https://github.com/SatyadevNtv/MOT}.


\section{Optimal transport problem formulations} \label{sec:motivation}
We consider the discrete case, where the marginals $\mu_1$ and $\mu_2$ of size $m$ and $n$, respectively. Hence, the vectors $\mu_1$ and $\mu_2$ lie on simplices $\Delta_{m}$ and $\Delta_n$, respectively, where $\Delta_{k}\coloneqq\{\bz\in\R_{+}^k|\sum_i \bz_i=1\}$. In this section, we briefly discuss the popular OT problems.

Problem (\ref{eqn:optimal-transport}) is equivalently written as 
\begin{equation}\label{eqn:discrete-optimal-transport}
   \minop_{\bGamma\in\Pi(\mu_1,\mu_2)} \langle \bGamma, \bC \rangle,
\end{equation}
where the transport plan $\bGamma$ is a matrix of size $m \times n$, $\bC$ is the {cost matrix} associated between $m$ points in $\calX$ and $n$ points in $\calY$, and $ \langle \bGamma, \bC \rangle = \trace(\bGamma ^\top \bC)$. The constraint set $\Pi(\mu_1, \mu_2)$ has the characterization
\begin{equation}\label{eqn:Pi}
    \Pi(\mu_1, \mu_2) \coloneqq \{ \bGamma \in \mathbb{R}^{m \times n} : \bGamma \geq 0, \bGamma \bone_n = \mu_1, {\rm \ and \ } \bGamma^\top \bone_m = \mu_2 \},
\end{equation}
where $\bone_m$ and $\bone_n$ are the column vectors of ones and of length $m$ and $n$, respectively.

Recent works \cite{paty19a,kolouri19a,deshpande19a,paty19a,jawanpuria2020efficient} have proposed robust variants of the Wasserstein distance that aim at maximizing the minimal transport cost over a set of ground cost functions. In particular, they solve the problem of the form
\begin{equation}\label{eqn:robust-optimal-transport}
   \minop_{\bGamma\in\Pi(\mu_1,\mu_2)} \maxop_{\bC \in \calC}  \ \langle \bGamma, \bC \rangle,
\end{equation}
where $\calC$ is the set of cost functions. In many applications, the solution of the inner maximization problem is either in closed form or easy to solve with a solver. Problem (\ref{eqn:robust-optimal-transport}) can, therefore, be treated as an optimization problem in $\bGamma$.

The Gromow-Wasserstein (GW) distance aims at learning optimal transport between metric spaces, i.e., given two measured similarity matrices $\bS_1$ and $\bS_2$ of size $m\times m$ and $n\times n$, respectively, the GW distance involves solving the problem 
\begin{equation}\label{eqn:general_GW}
    \minop_{\bGamma\in\Pi(\mu_1,\mu_2)} \sum_{ijkl} \ell({\bS}_{1,ik}, {\bS}_{2,jl})\bGamma_{ij} \bGamma_{kl},
\end{equation}
where $\ell$ is any well-defined loss function. Specifically, when $\ell$ is the squared Frobenius norm loss, the optimization problem (\ref{eqn:general_GW}) boils down to 
\begin{equation*}\label{eqn:GW}
    \minop_{\bGamma\in\Pi(\mu_1,\mu_2)} - \langle \bGamma^\top \bS_1 \bGamma, \bS_2 \rangle.
\end{equation*}

Overall, Problems (\ref{eqn:discrete-optimal-transport}), (\ref{eqn:robust-optimal-transport}), and (\ref{eqn:general_GW}) can be written as 
\begin{equation}\label{eqn:constrained_OT}
    \minop_{\bGamma\in\Pi(\mu_1,\mu_2)} f(\bGamma),
\end{equation}
where $f$ is a general (non-linear) OT objective function.

A popular way to solve the constrained optimization problem (\ref{eqn:constrained_OT}) is with the Frank-Wolfe (FW) algorithm \cite{jaggi2013revisiting,ferradans14a,paty19a}, which is also known as the conditional gradient algorithm. It requires solving a constrained linear minimization sub-problem (LMO) at every iteration. The LMO step boils down to solving a standard optimal transport problem (\ref{eqn:optimal-transport}). 
When regularized with an entropy regularization term, the LMO step (at every iteration) boils down to solving the problem 
\begin{equation}\label{eqn:lmo}
    \minop_{\widetilde{\bGamma}  \in \Pi(\mu_1, \mu_2)} \langle \widetilde{\bGamma}, \nabla f(\bGamma_0) \rangle + \epsilon \sum_{ij} \widetilde{\bGamma}_{ij} \rm{log}(\widetilde{\bGamma}_{ij}),
\end{equation}
where $\bGamma_0$ is the current iterate and $\nabla f $ is the partial derivative of $f$. Here, $\epsilon$ is the amount of entropy regularization. The LMO (\ref{eqn:lmo}) 
admits a computationally efficient solution using the Sinkhorn algorithm \cite{cuturi13a}. Even though the FW algorithm is appealing because of its simplicity, tuning of the regularization parameter $\epsilon$ introduces a practical challenge. It should be noted that the use of entropy regularization implies that the learned $\bGamma > 0$ (has positive values), i.e., strictly positive joint probabilities are learned.

\section{Optimal transport as a manifold optimization problem} \label{sec:manifold_optimization}


As an alternative approach to the FW algorithm, we discuss the manifold view of approaching optimal transport problems which does not require explicit regularization, and is yet versatile enough to be used in various settings. The basis of this is the fact that $\Pi(\mu_1,\mu_2)$ has a {differentiable} manifold structure provided we enforce $\bGamma > 0$ , i.e., the boundary is ``removed'' \cite{douik2019manifold,shi19a}. 
This implies that we also learn only strictly positive joint distributions via the Riemannian optimization framework as is the case of solving OT problems with entropy regularization (e.g., with the FW algorithm). 
We denote the interior of $\Pi(\mu_1, \mu_2)$ by $\mathcal{M}(\mu_1,\mu_2)$, i.e.,
\begin{equation}\label{eqn:set}
\mathcal{M}(\mu_1,\mu_2) \coloneqq \{ \bGamma \in \mathbb{R}^{m \times n} : \bGamma > 0, \bGamma \bone_n = \mu_1, {\rm \ and \ } \bGamma^\top \bone_m = \mu_2 \},
\end{equation}
where $\bone_m$ and $\bone_n$ are the column vectors of ones and of length $m$ and $n$, respectively.

We define a smooth inner product, also called a {metric}, on $\mathcal{M}(\mu_1,\mu_2)$. Once a metric is defined, the manifold $\mathcal{M}(\mu_1,\mu_2)$ has the structure of a Riemannian manifold (and the metric is called the Riemannian metric). In particular, it has the structure of a Riemannian {submanifold} of $\mathbb{R}^{m \times n}$. Consequently, the general (non-linear) optimal transport problem is seen as an optimization problem over the non-linear Riemannian submanifold $\mathcal{M}(\mu_1,\mu_2)$, i.e.,
\begin{equation}\label{eqn:problem_manifold}
    \minop_{\bGamma \in \mathcal{M}(\mu_1, \mu_2)} \quad f(\bGamma),
\end{equation}
where $f$ is the optimal transport objective function.

Below we brief describe various optimization-related ingredients that enable first and second order optimization for (\ref{eqn:problem_manifold}). Most of the material presented below are borrowed from \cite{douik2019manifold,shi19a,sun2015heterogeneous}.

\subsection{Metric}\label{sec:metric}

$\mathcal{M}(\mu_1,\mu_2)$ is endowed with the {Fisher} information metric, as the Riemannian metric, defined as \cite{douik2019manifold,shi19a,sun2015heterogeneous}
\begin{equation}\label{eqn:Riemannian_metric}
    \begin{array}{ll}
         g_{\bGamma}(\eta_{\bGamma}, \xi_{\bGamma}) \coloneqq 
         \trace((\eta_{\bGamma} \odot \xi_{\bGamma}) \oslash \bGamma),
    \end{array}
\end{equation}
where $\eta_{\bGamma}$ are $ \xi_{\bGamma}$ tangent vectors belonging to the tangent space $T_{\bGamma} \mathcal{M}(\mu_1,\mu_2) \coloneqq \{ \zeta_{\bGamma} \in \mathbb{R}^{m\times n} : \zeta_{\bGamma} \bone = \bzero_m, \zeta_{\bGamma}^\top \bone =  \bzero_n \}$ at $\bGamma$. Here, $\odot$ and $\oslash$ are element-wise matrix multiplication and division, respectively. 

The Riemannian metric (\ref{eqn:Riemannian_metric}) scales the boundary of $\mathcal{M}(\mu_1,\mu_2)$ to ``infinity'', i.e., as $\bGamma_{ij}$ tends to zero, the inner product tends to infinity, thereby endowing $\mathcal{M}(\mu_1,\mu_2)$ with a smooth {differentiable} structure. It is interesting to note that the Fisher metric (\ref{eqn:Riemannian_metric}) is same as the popular Bures-Wasserstein metric (defined for the manifold of symmetric positive definite matrices) when restricted to positive scalars \cite{bhatia2019bures,malago2018wasserstein}.

As the metric (\ref{eqn:Riemannian_metric}) involves only element-wise operations of matrices, the computation costs $O(mn)$.

\subsection{Linear projector onto tangent space} \label{sec:projector}

An important ingredient is the linear projection operator that projects a vector in the ambient space $\mathbb{R}^{m \times n}$ onto $T_{\bGamma} \mathcal{M}(\mu_1, \mu_2)$.

If $\bZ \in \mathbb{R}^{m \times n}$, then the projection operator ${\rm Proj}_{\bGamma}(\bZ): \mathbb{R}^{m \times n} \rightarrow T_{\bGamma} \mathcal{M}(\mu_1, \mu_2)$ is defined as 
\begin{equation}\label{eqn:projection}
{\rm Proj}_{\bGamma}(\bZ) = \bZ - ({\alpha}\bone_{n}^\top + \bone_{m} \beta^\top)\odot{\bGamma}   , 
\end{equation}
where the unique $\alpha$ and $\beta$ values are obtained by solving the linear system
\begin{equation}\label{eqn:linearsolve}
\begin{array}{ll}
\alpha \odot \mu_1 + \bGamma \beta =  \bZ \bone_n \\ 
 \beta\odot \mu_2  + \bGamma^\top \alpha  = \bZ^\top \bone_m.
\end{array}
\end{equation}

The standard way to solve (\ref{eqn:linearsolve}) is rewrite it as ``$\bA x = b$'' form and optimal the solution corresponds to ``$\bA^ {-1} b$''. Although simple, this is computationally costly. An appealing alternative is to solve  (\ref{eqn:linearsolve}) with an iterative solver, where each iteration costs $O(mn)$. In practice, we only need a fix number of such iterations, making the iterative approach more attractive.

\subsection{Retraction}
Given a search direction in the tangent space, the {retraction} operator maps it to an element of the manifold \cite{absil2009optimization}. It provides a convenient way to move on the manifold. On $\mathcal{M}(\mu_1, \mu_2)$, the retraction has a well-defined expression using the Sinkhorn algorithm \cite{douik2019manifold}, i.e., 
\begin{equation}\label{eqn:retraction}
    R_{\bGamma}({\xi_{\bGamma}}) = {\rm Sinkhorn}(\bGamma \odot{\rm exp}(\xi_{\bGamma} \oslash \bGamma)),
\end{equation}
where ${\rm exp}$ is element-wise exponential operator and ${\rm Sinkhorn}$ implies the Sinkhorn algorithm. It should be noted that $R_{\bGamma}({\xi_{\bGamma}}) \in \mathcal{M}(\mu_1, \mu_2)$. 

It should be noted that the Sinkhorn algorithm \cite{douik2019manifold} here comes naturally in the manifold optimization framework motivated by geometry of the space and is independent of the objective function at hand. This is in contrast to traditional OT solution approaches like \cite{cuturi13a}, where the Sinkhorn algorithm is used in the context of a linear objective function only.

The element-wise exponential operation costs $O(mn)$. Each iteration inside the Sinkhorn algorithm results in element-wise operations costing $O(mn)$, and we need a fix number of Sinkhorn iterations in practice. Overall, the retraction operation scales as $O(mn)$.

\subsection{Computations of Riemannian gradient and Hessian}
First and second other algorithms require the notions of Riemannian {gradient} (the steepest descent direction on the manifold) and {Hessian} (the covariant derivative of the Riemannian gradient). These admit simple formulas in terms of the partial derivative $\nabla_{
\bGamma}f$ and its directional derivative ${\rm D} \nabla_{\bGamma }f [\xi_{\bGamma}]$ along a search direction $\xi_{\bGamma} \in T_{\bGamma} \mathcal{M}(\mu_1, \mu_2)$, respectively \cite{douik2019manifold,sun2015heterogeneous}. In particular, 
\begin{equation}
    \begin{array}{lll}
         \text{ the\ Riemannian\ gradient\  grad}_{\bGamma} f& =& {\rm Proj}_{\bGamma}(\bGamma\odot \nabla_{\bGamma }f)  {\rm \ and }\\
         \text{the\ Riemannian\ Hessian\ Hess}_{\bGamma} f [\xi_{\bGamma}]& =& {\rm Proj}_{\bGamma}( {\rm D} {\rm grad}_{\bGamma} f [\xi_{\bGamma}] - \frac{1}{2} {\rm grad}_{\bGamma} f\odot \xi_{\bGamma} \oslash \bGamma ),
    \end{array}
\end{equation}
where $ {\rm D} {\rm grad}_{\bGamma} f [\xi_{\bGamma}]$ is the directional derivative of the Riemannian gradient $\grad_{\bGamma }f$ along the search direction $\xi_{\bGamma}$. $ {\rm D} {\rm grad}_{\bGamma} f [\xi_{\bGamma}]$ can be written in terms of $\nabla_{
\bGamma}f$ and ${\rm D} \nabla_{\bGamma }f [\xi_{\bGamma}]$.

It should be noted that $\nabla_{
\bGamma}f$ and ${\rm D} \nabla_{\bGamma }f [\xi_{\bGamma}]$ easy to compute as they only involve the optimal transport objective function $f$ (we provide explicit formulas for various optimal transport formulations of interest). The costs of computation of Riemannian gradient and Hessian are dominated by the costs of $\nabla_{
\bGamma}f$ and ${\rm D} \nabla_{\bGamma }f [\xi_{\bGamma}]$.


\section{Extensions} \label{sec:extension}
In this section, we consider three extensions to the optimal transport problem and how to computationally approach them in the discussed manifold optimization framework. 



\subsection{Dealing with sparse optimal transport map}
In this scenario, we look at how take explicit sparsity constraints in $\bGamma$ into account. From a modeling perspective, it is equivalent to considering the set
\begin{equation}\label{eqn:sparse_set}
\mathcal{M}_{\Omega}(\mu_1,\mu_2) \coloneqq \{ \bGamma^{m \times n} : \bGamma > 0, \bGamma \bone_n = \mu_1,  \bGamma^\top \bone_m = \mu_2, {\rm \ and \ }  \bGamma_{\Omega} = 0\},
\end{equation}
where $\Omega$ is a subset of indices for which the values in $\bGamma$ are explicitly known to be zero. It should be noted that the set $\mathcal{M}_{\Omega}(\mu_1,\mu_2)$ (\ref{eqn:sparse_set}) is a restricted version of $\mathcal{M}(\mu_1,\mu_2)$ defined in (\ref{eqn:set}). 

It is straightforward to show that the set $\mathcal{M}_{\Omega}(\mu_1,\mu_2)$ is a Riemannian submanifold of $\mathbb{R}^{m\times n}$ by ensuring that the systems of linear equations, arising from linear equality constraints, in (\ref{eqn:sparse_set}) is {underdetermined}. Consequently, all the optimization-related ingredients in Sections \ref{sec:metric} and \ref{sec:projector} follow through by explicitly maintaining a sparse $\bGamma$. To reuse the Sinkhorn algorithm in the retraction defined in (\ref{eqn:retraction}), we need to additionally ensure that the sparsity pattern $\Omega$ leads to total support for the term $\bGamma \odot{\rm exp}(\xi_{\bGamma} \oslash \bGamma)$ \cite{knight2008sinkhorn,cariello2019sinkhorn}. An interesting sparsity pattern that can be readily tackled is the block diagonal sparsity structure.

\subsection{Learning of many transport plans simultaneously}

Another useful extension is when we need to solve $k$ optimal transport problems simultaneously, i.e.,
\begin{equation}\label{eqn:k_problem_manifold}
    \minop_{
    \begin{array}{lll}
    \bGamma_1 \in \mathcal{M}(\mu_1, \mu_2) \\
    \bGamma_2 \in \mathcal{M}(\mu_1, \mu_2) \\
    \vdots \\
    \bGamma_k \in \mathcal{M}(\mu_1, \mu_2)
    \end{array}
    } \quad f(\bGamma_1, \bGamma_2, \ldots,\bGamma_k).
\end{equation}
An approach for (\ref{eqn:k_problem_manifold}) is to construct a tensor $\mathbfcal{T} \coloneqq [\bGamma_1, \bGamma_2, \ldots, \bGamma_k]$ of size $m\times n\times k$ and update optimize $\mathbfcal{T}$ directly. To this end, all the ingredients proposed in Section \ref{sec:manifold_optimization} are easily tensorizable. The main advantage of the tensorization comes in computing the projection (\ref{eqn:projection})  and retraction (\ref{eqn:retraction}) operations.

\section{Experiments} \label{sec:experiments}
We consider the following non-linear OT problems for our empirical evaluations: a) Gromov-Wasserstein (GW) OT problem~\cite{peyre16a}, and b) Co-optimal transport (COOT) problem~\cite{titouan2020co}. 

\textbf{Gromov-Wasserstein OT problem}: we consider the GW optimization problem (\ref{eqn:general_GW}) with the squared Frobenius norm loss. The POT library~\cite{flamary2021pot} provides an efficient solution of the GW problem (henceforth termed as POT-GW). 


\textbf{Co-optimal transport problem}: this extends the traditional optimal transport setting to heterogeneous domains \cite{titouan2020co}. In particular, it involves learning two transport plan matrices, i.e.,
\begin{equation*}
\minop_{
    \begin{array}{ll}
    \bGamma_1 \in \mathcal{M}(\mu_1, \mu_2) \\
    \bGamma_2 \in \mathcal{M}(\nu_1, \nu_2)
    \end{array}
    } \ell(\bX_{ik}, \bZ_{jl}) \bGamma_{1,ij} \bGamma_{2,kl},
\end{equation*}
where $\ell$ is any well defined loss between two scalars, $\bX \in \mathbb{R}^{m\times d_1}$ and $\bZ \in \mathbb{R}^{n\times d_2}$ are given data matrices from different heterogeneous domains, and $\bGamma_1$ and $\bGamma_2$ are transport plans of sizes $m\times n$ and $d_1\times d_2$, respectively. Here, $\bGamma_1$ is the transport plan between the data samples and $\bGamma_2$ is the transport plan between the different features. Titouan et al \cite{titouan2020co} propose an alternate minimization (AM) algorithm for the COOT problem. We consider the case when $\ell$ is the square loss.

\textbf{Baselines}: As discussed, MOT is a generic manifold optimization framework for non-linear OT problems. Existing works \cite{courty17b,paty19a,jawanpuria2020efficient} have employed the FW algorithm in such settings. In fact, both the AM algorithm (for COOT) and the POT-GW algorithm (for GW) may also be viewed as a limiting case of the FW algorithm (with per-iteration step size fixed as $1$). Hence, we study the convergence quality of MOT and compare it with the FW algorithm, the AM algorithm (for the COOT setting), and the POT-GW algorithm (for the GW setting). 

\textbf{Experimental Setup}: For both the COOT and GW settings, we consider the MNIST dataset as the source domain and the USPS dataset as the target domain~\cite{titouan2020co}. 
In the COOT experiments, we randomly sample $2\,141$ points for the source and $7\,291$ points for the target domains. In GW experiments, we randomly sample $1\,000$ points for both the source and the target domains. The source and the target marginals in both the setups are generated as follows: the discrete marginal distribution for $k$ points (samples/features) is randomly sampled from the $k-1$ dimensional simplex. We report the OT cost versus time plots over different random seeds (leading to different marginals and samples). 

\begin{figure*}[t]
\begin{center}
\begin{tabular}{ccc}
\begin{minipage}{0.3\hsize}
\begin{center}
\includegraphics[width=\hsize]{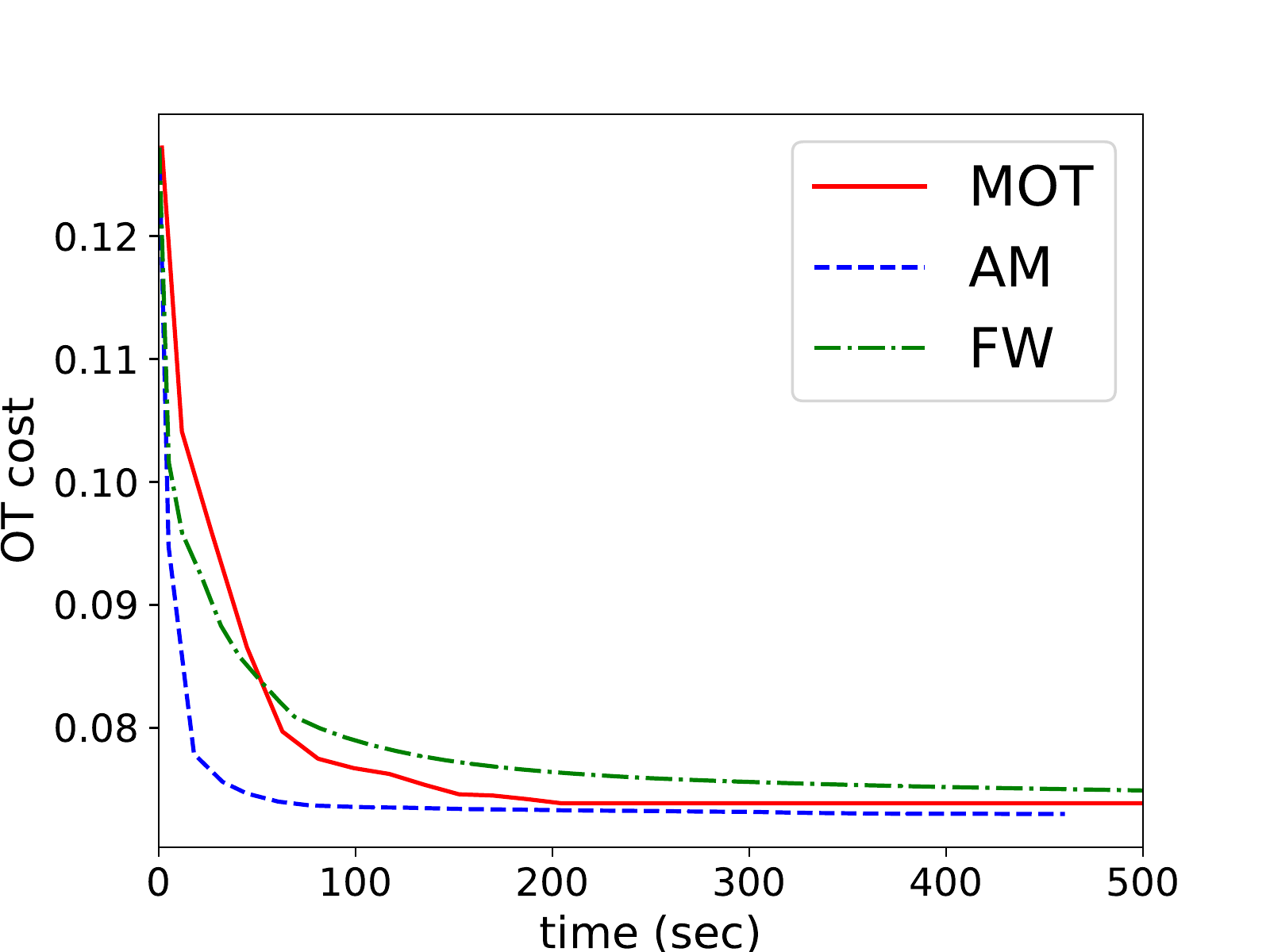}\\
{(a)}
\end{center}
\end{minipage}
&
\begin{minipage}{0.3\hsize}
\begin{center}
\includegraphics[width=\hsize]{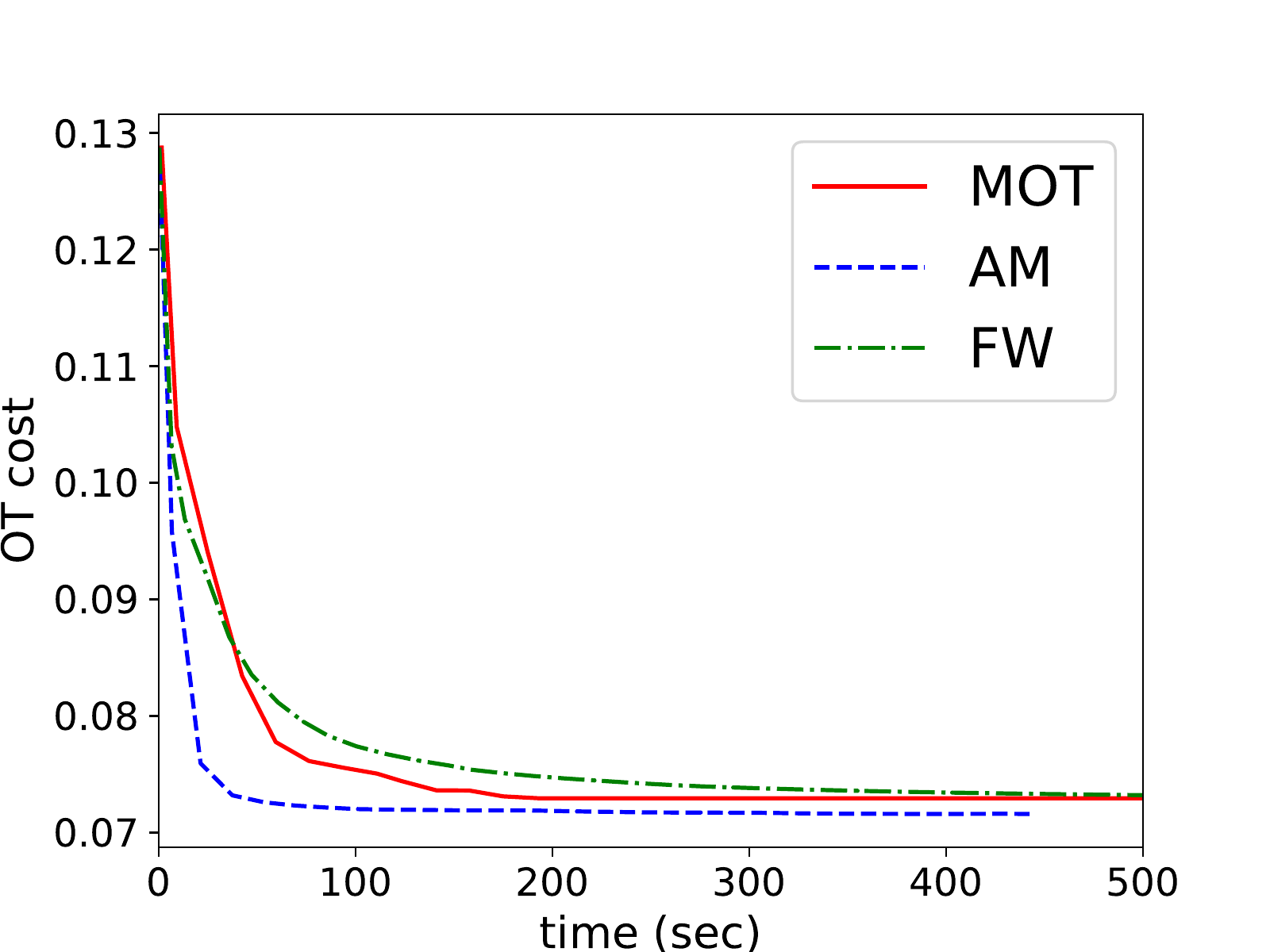}\\
{(b)}
\end{center}
\end{minipage}
&
\begin{minipage}{0.3\hsize}
\begin{center}
\includegraphics[width=\hsize]{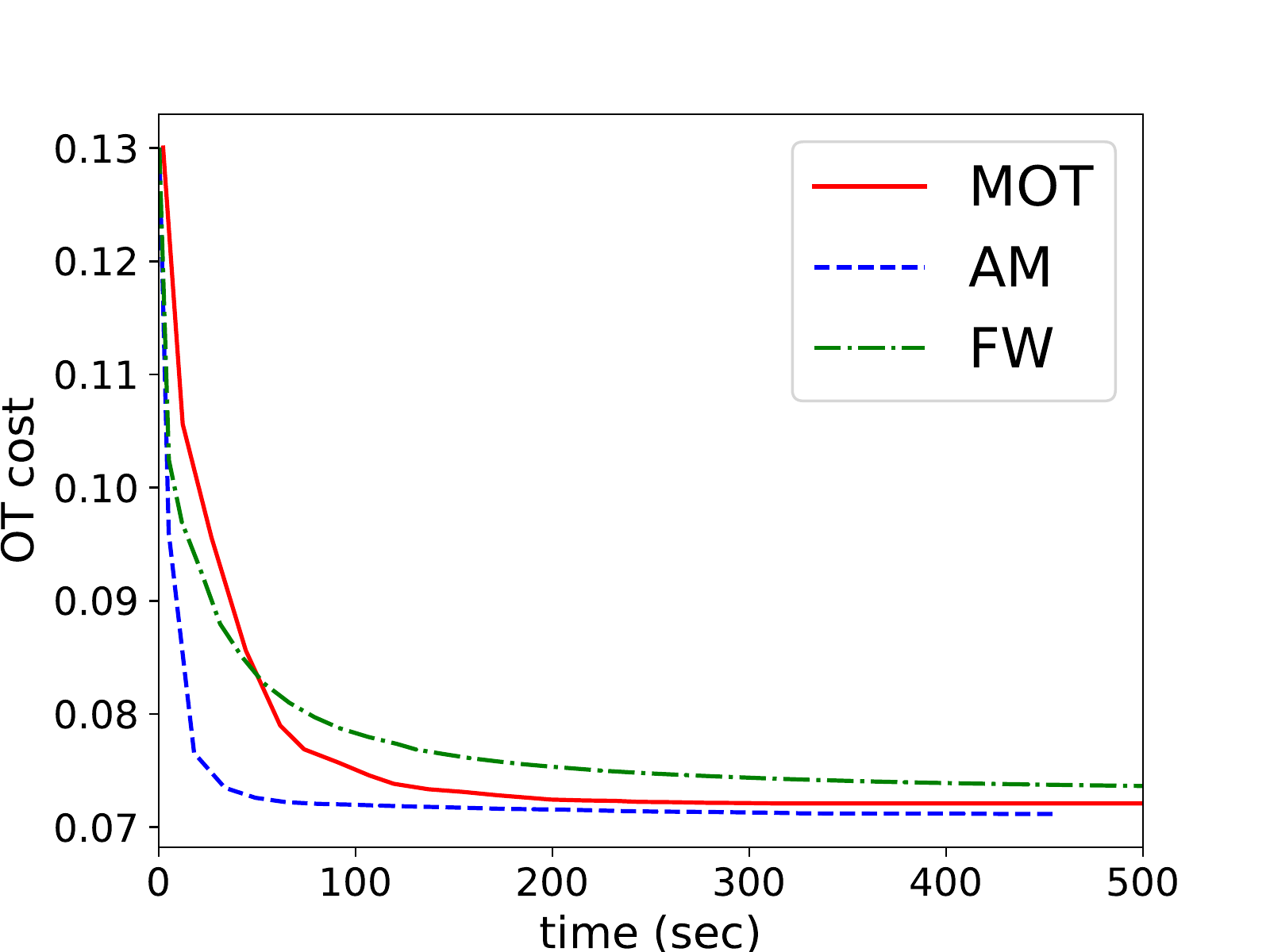}\\
{(c)}
\end{center}
\end{minipage}
\end{tabular}
\caption{The evolution of OT cost with time in three different COOT problems. 
}
\label{fig:COOT}
\end{center}
\end{figure*}
\begin{figure*}[t]
\begin{center}
\begin{tabular}{ccc}
\begin{minipage}{0.3\hsize}
\begin{center}
\includegraphics[width=\hsize]{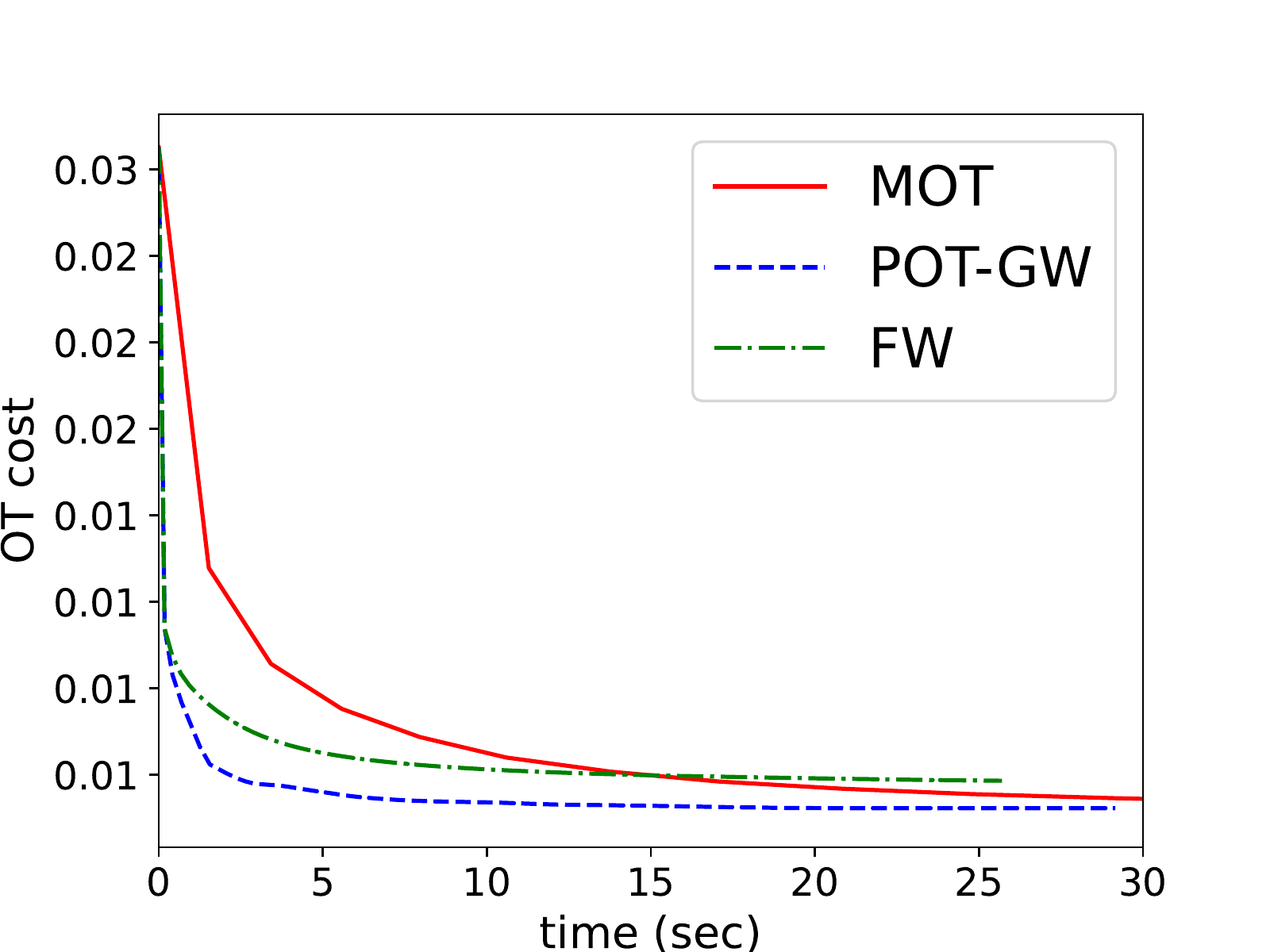}\\
{(a)}
\end{center}
\end{minipage}
&
\begin{minipage}{0.3\hsize}
\begin{center}
\includegraphics[width=\hsize]{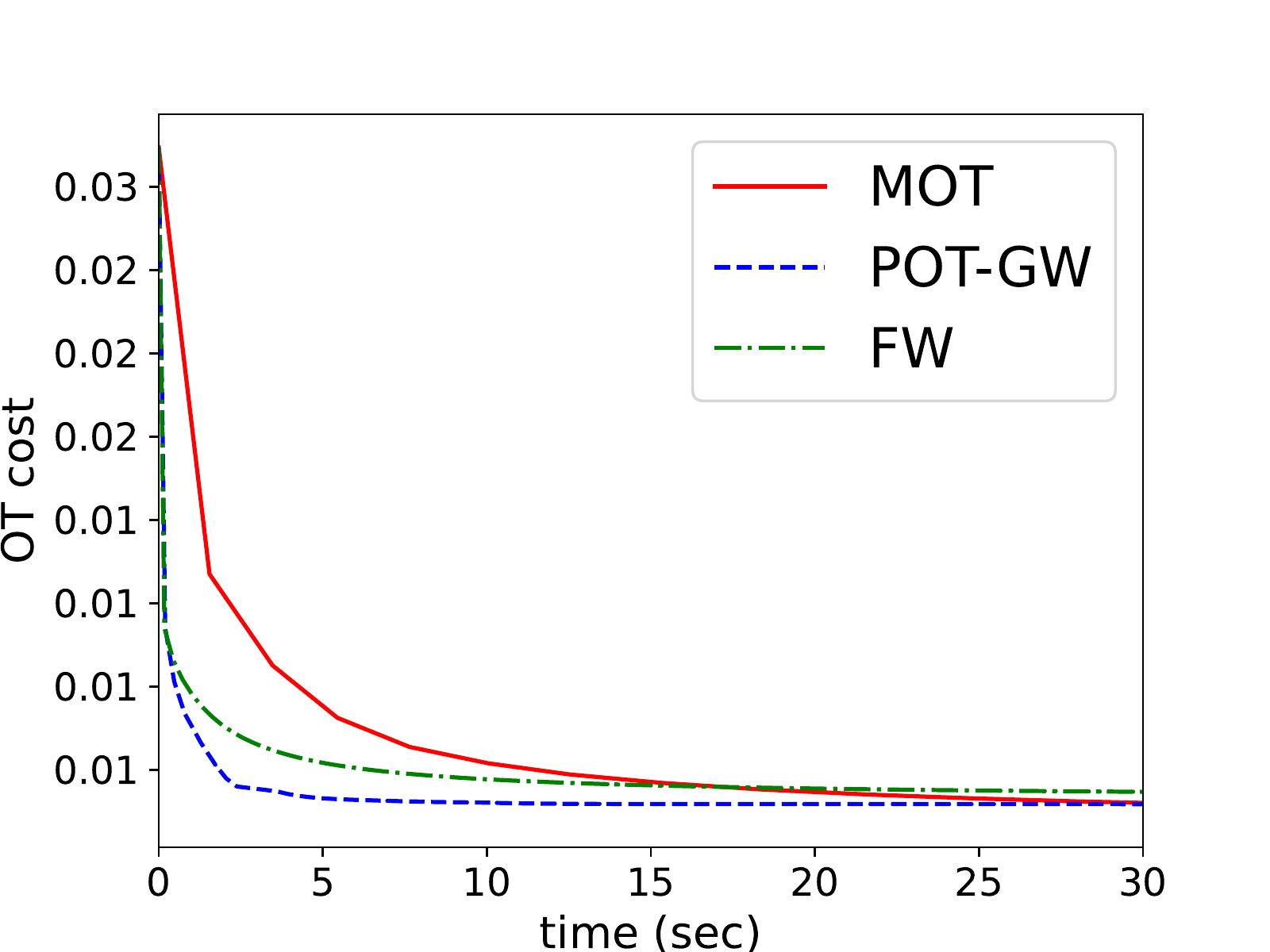}\\
{(b)}
\end{center}
\end{minipage}
&
\begin{minipage}{0.3\hsize}
\begin{center}
\includegraphics[width=\hsize]{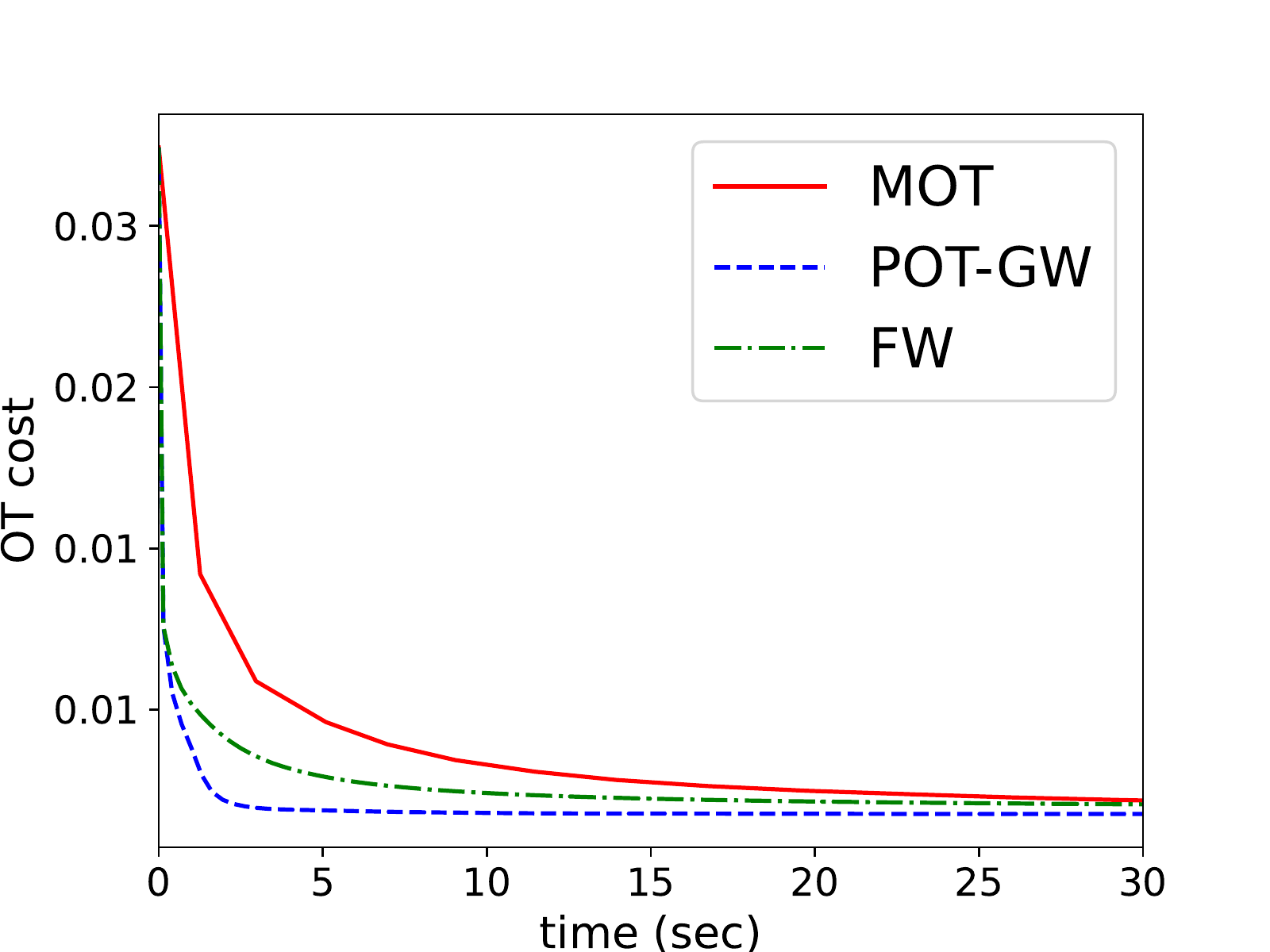}\\
{(c)}
\end{center}
\end{minipage}
\end{tabular}
\caption{The evolution of OT cost with time in three different GW problems.
}
\label{fig:GW}
\end{center}
\end{figure*}

\textbf{Results}: The plots for COOT and GW experiments are presented in Figure~\ref{fig:COOT} and Figure~\ref{fig:GW}, respectively. We observe that the manifold optimal transport algorithm shows a good quality convergence and is comparable to the baselines while maintaining full generality. 


\section{Conclusion}
We have discussed a manifold approach to solving various optimal transport (OT) problems. This gives an alternative viewpoint for solving non-linear OT problems, which has traditionally been motivated by the work of Cuturi \cite{cuturi13a} via entropic regularization. In particular, the Frank-Wolfe algorithm and its variants (with iterates involving the Sinkhorn algorithm) have been a popular choice for non-linear OT problems. However, they entail the use of entropic regularization. The manifold framework, on the other hand, offers another principled route for tackling general non-linear OT problems by exploiting the space of joint probability distributions. We observe that the manifold framework allows usage of non-entropic regularization schemes on the transport plan but still take advantage of the celebrated Sinkhorn iterations for an efficient solution.  
Exploiting the manifold viewpoint, the optimization-related ingredients have easy to implement expressions, which we make available as part of the MOT repository. 
